# A Transformer Model for Boundary Detection in Continuous Sign Language

Razieh Rastgoo[1], Kourosh Kiani [1]*, Sergio Escalera[2]

**Abstract.** Sign Language Recognition (SLR) has garnered significant attention from researchers in recent years, particularly the intricate domain of Continuous Sign Language Recognition (CSLR), which presents heightened complexity compared to Isolated Sign Language Recognition (ISLR). One of the prominent challenges in CSLR pertains to accurately detecting the boundaries of isolated signs within a continuous video stream. Additionally, the reliance on handcrafted features in existing models poses a challenge to achieving optimal accuracy. To surmount these challenges, we propose a novel approach utilizing a Transformer-based model. Unlike traditional models, our approach focuses on enhancing accuracy while eliminating the need for handcrafted features. The Transformer model is employed for both ISLR and CSLR. The training process involves using isolated sign videos, where hand keypoint features extracted from the input video are enriched using the Transformer model. Subsequently, these enriched features are forwarded to the final classification layer. The trained model, coupled with a post-processing method, is then applied to detect isolated sign boundaries within continuous sign videos. The evaluation of our model is conducted on two distinct datasets, including both continuous signs and their corresponding isolated signs, demonstrates promising results.
**Keywords:** Sign Language Recognition (SLR), Continuous Sign Language Recognition (CSLR), Isolated Sign Language Recognition (ISLR), Transformer, Hand pose, Boundary detection.

# 1. Introduction

Sign language, as a visual language, is a communication proficiency for persons with speech and hearing impairments [1]. Generally, there are five fundamental parameters in sign language: hand shape, orientation, movement, location, and components such as mouth shape and eyebrow movements [2]. Sign language can be categorized into two groups: isolated and continuous sign language. These groups are concentrated on an image/video of only one sign word and a video of a sign sentence, as a model input, respectively. Previous works are mostly focused on Isolated Sign Language Recognition (ISLR) [1]. Although, some challenges, such as hand occlusions, fast movement, lack of large and diverse datasets, background complexity, and illumination conditions, are still remained in ISLR. In addition to the challenges, other challenges, such as the boundary detection of the isolated signs as well as the different lengths of the sentences, need to be discussed and solved in Continuous Sign Language Recognition (CSLR) [3]. Different models have been conducted to sign language translation of the aforementioned groups, ISLR and CSLR, to the text/voice, which could facilitate communication between the Deaf and other people in society. However, those who do not know sign language usually undervalue or reject persons with such an impairment. Therefore, this makes it inevitable to propose a two-way translation application for sign language to text/voice conversion and vis versa. Focusing on the one-way of such a system, sign language to text/voice [4], we propose a model to tackle the challenge of boundary detection in CSLR using the recent advancement in Deep Learning for Sign Language Recognition (SLR) improvement to fill the gap between the hearing and Deaf communities.

[1] Razieh Rastgoo
Electrical and Computer Engineering Department, Semnan University, Semnan, 3513119111, Iran
E-mail: rrastgoo@semnan.ac.ir, ORCID: 0000-0001-7963-9461

[1*] Kourosh Kiani (*Corresponding Author)
Fax: +98-23-33-654123
Electrical and Computer Engineering Department, Semnan University, Semnan, 3513119111, Iran
E-mail: Kourosh.kiani@semnan.ac.ir, ORCID: 0000-0001-6582-8691

[2] Sergio Escalera
Department of Mathematics and Informatics, Universität de Barcelona, and Computer Vision Center, Barcelona, Spain
E-mail: sescalera@ub.edu, ORCID: 0000-0003-0617-8873

Recent advances in Deep Learning have encouraged researchers to develop different applications handling various vision tasks [5]. Many thanks to the well-known companies for developing efficient devices and applications, these products are often voice-based and unavailable for free in Deaf community. With the aim of providing an automatic, user-friendly, and freely available sign translation application for establishing a bidirectional communication between a Deaf user and a digital assistant/hearing people in real-time, a fast and accurate model capable of working in parallel is needed. Considering this as well as the superior performance of Transformer models in many tasks related to Natural Language Processing (NLP) and Computer Vision (CV) [6], we propose a model for accurate boundary detection of isolated signs in continuous sign videos to translate them into text/voice. The proposed model obtains promising results in continuous sign language recognition as a challenging research area.

In summary, we list our contributions:

- **Model**: We propose a pose-based model using recent advances in Transformer models for boundary detection of the isolated signs in the continuous video streams. While the proposed model is trained on the ISLR, it will be evaluated on continuous video streams without any pre-processing.
- **Multi-tasking:** While the proposed model is trained on isolated sign videos, it will be used for continuous sign videos. So, both the ISLR and CSLR are handled by only one model.
- **Performance**: The proposed model obtains promising results on two datasets.

In the remainder of this paper, Section 2 presents a brief introduction to recent works in SLR. The proposed model as well as the experimental results will be presented in sections 3 and 4, respectively. Finally, a discussion along with the conclusion on the work with a roadmap for future work will be shown in section 5.

# 2. Related works

In this section, recent deep learning-based works in CSLR are discussed in short. Compared to ISLR, there are fewer works in CSLR [1]. Generally, most of the previous works in CSLR are focused on the visual characteristics of signers in isolated/continuous video streams to make the accurate recognizer model [3]. Thanks to researchers for developing these models in ISLR and CSLR with promising results, there are still some challenges that need to be addressed. One of these challenges is detecting the isolated sign boundaries in a continuous sign video stream, which is complex due to the different sequential patterns in the unseen continuous signs. One solution to this challenge is pre-training and using an iterative training mechanism [6], which greatly lengthens the training process. Generally, using the pre-training mechanism to transfer the obtained knowledge in ISLR into CSLR is a useful solution for this challenge. However, the model complexity as well as the parallel computing of the model need to be addressed in these models to make a room for developing real-time applications. In this way, we categorize the recent works in SLR as follows:

- **Pre-trained models:** In this category, the pre-trained models are used to enhance the model performance in SLR. To this end, various methodologies have been developed, such as using the knowledge transferred from the bigger dataset of British Sign Language (BSL) to the target model [7] as well as the first 20 layers of the pre-trained InceptionV3 for static Swedish Sign Language (SSL) recognition [8]. In addition to these, different transfer learning configurations of the AlexNet model have been used for Chinese SLR [9]. In another work, the obtained knowledge from the isolated word sign dataset is transferred to the continuous sign video samples [10]. Relying on the transfer learning approach in this category, in this paper, we propose to transfer the obtained knowledge from the trained model on the isolated signs into the continuous signs, relying on large datasets with a large number of samples in each class.
- **Non-Pre-trained models:** The models in this category do not use the pre-training or transfer learning mechanism to benefit from the knowledge of ISLR in CSLR. Instead, different feature extractor models, especially deep learning-based models, are used in these models. For instance, Mocialov et al. proposed a model using a heuristic-based method for the segmentation of the video stream combined with stacked LSTMs. This model aims to recognize the epenthesis in the continuous video and apply a classifier to the detected sign glosses. Results on a dataset, including a sub-set of the NGT corpus containing approximately 100 participants telling stories or having discussions with other Dutch sign language users, have led to the accuracy of over 80% and 95% on the continuous and isolated video signs, respectively [11]. Although the results are promising, their model is not robust to the signer-independent condition [2]. Papastratis et al. introduced a deep generative model, entitled Sign Language Recognition Generative Adversarial Network (SLRGAN), for CSLR. In addition, a transformer network is used to produce a natural language text from the sign language glosses. Results on the RWTH-Phoenix-Weather-2014, the Chinese Sign Language (CSL), and Greek Sign Language (GSL) Signer Independent (SI) datasets show the word error rates of 23.4%, 2.1%, and 2.26% on these datasets, respectively [12]. Koishybay et al. developed a Deep Learning-based model for CSLR using iterative Gloss Recognition (GR) fine-tuning. In this way, a spatiotemporal feature-extraction model has been embedded in the model for gloss features segmentation. Furthermore, a BiLSTM Network along with the CTC is used as a sequence learning model. The feature extractor model is iteratively fine-tuned to segment the sign glosses with alignments from the end2end model. Results on the RWTH-PHOENIX-Weather-2014 dataset show 34.4 Word Error Rate (WER) for test data [13]. Cui et al. proposed a model using deep convolutional neural networks with stacked temporal fusion layers and bi-directional recurrent neural networks for CSLR. In this way, their proposed model is trained for alignment proposal to enrich the feature extraction mechanism. The training process is iteratively continued using two input modalities, RGB images and optical flow. Results on two continuous sign language datasets, RWTH-PHOENIX-Weather multi-signer 2014 dataset and SIGNUM signer-dependent dataset, confirm the superiority of the model compared to the state-of-the-art models by a margin improvement of more than 15% on both datasets [14]. Zuo and Mak introduced two auxiliary constraints to improve the CSLR performance. These constraints aim to improve the visual module as well as the language module in the model. In this way, a pose-based spatial attention module is used in the visual module for more concentration on the informative regions. In addition to this, making consistency between the visual and language modules is necessary for the representation improvement of both features. Results on three benchmarks, PHOENIX-2014, PHOENIX-2014-T, and CSL, show that the performance of this model outperforms state-of-the-art models for CSLR [15].

# 3. Proposed model

Recently, Transformer models have outperformed the previous Recurrent Neural Network (RNN)-based models in dealing with long-range dependencies [16]. A Transformer is a deep learning-based model that adopts the self-attention mechanism, differentially weighting the significance of each part of the input data. Transformers are increasingly the model of choice for NLP problems, however, the Vision Transform [17] has borrowed the idea of the Transformer in CV. Relying on the self-attention mechanism, the Transformer model prevents the recurrent structure by reducing the maximum length of network signals' traveling paths [16]. Inspired by NLP successes, multiple works try combining Convolutional Neural Network (CNN)-like architectures with self-attention [18-19], and some replace the convolutions entirely [20-21]. The latter models, while theoretically efficient, have not yet been scaled effectively on modern hardware accelerators due to the use of specialized attention patterns. Therefore, in large-scale image recognition, classic ResNet-like architectures are still state-of-the-art [22-24]. Considering the recently introduced models for coping with the challenge of boundary detection in a continuous sign video [3, 25], we propose a Transformer-based model for accuracy improvement as well as parallel computing and not dealing with the handcrafted features in the model. To this end, the proposed model is trained on the isolated sign videos using the 3D hand keypoint features obtained from the input sign video and fed to the Transformer model to enrich the features. After that, these rich features are fed to the final classification layer. The trained model along with the post-processing method is employed to detect the isolated sign boundaries in the continuous sign video. An overview of the proposed model for the training phase using the isolated signs is shown in the Fig. 1. Details of the 3D hand keypoints used in the model can be found in Fig. 2. Relying on the trained model on the isolated signs as well as the post-processing approach, the proposed model for the evaluation phase using the continuous sign videos is shown in Fig. 3. The proposed model includes two main steps, which are explained in details in the following:

- **Pre-Processing and training the model**: In this step, we aim to use the features that are automatically obtained from the deep model. In this way, the dataset is pre-processed in order to be used for model training and evaluation. Equalizing the frame numbers in each isolated video sign as well as the frame size, hand detection as well as 3D hand keypoint estimation using the OpenPose [26] in each frame, are done in a pre-processing step. After that, the 3D hand keypoint features of two hands are concatenated and fed to the Transformer model to enrich the features. A classification layer is stacked on the output features obtained from the Transformer for the final classification. The proposed model is trained on the isolated sign videos and used in the next step.
- **Post-Processing the predictions**: This step is only used in the evaluation phase, which the trained model is applied to the continuous sign videos. More specifically, a sliding window, including a predefined frame number, with stride one is applied to the continuous sign videos to detect the isolated signs. To this end, the 3D hand keypoints corresponding to each frame in the sliding window are fed to the Transformer model to obtain richer features. The concatenated features of all frames in a sliding window are input to a Fully Connected (FC) layer along with a Softmax layer for the sign classification. The recognized signs with recognition accuracy higher than a predefined threshold, which is 0.51 in our experiments, are accepted as the final recognized isolated signs. More details of the proposed pre-processing can be found in [3].

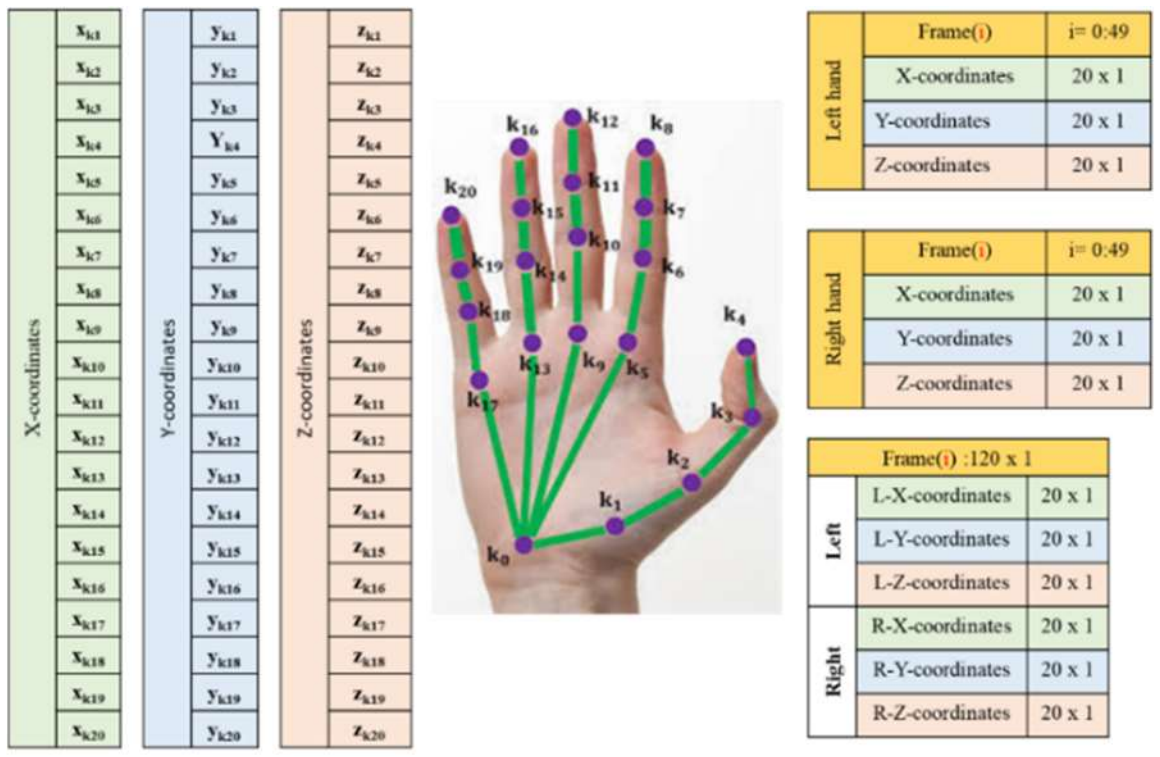

Fig. 1: Details of the 3D keypoints used in the proposed model.

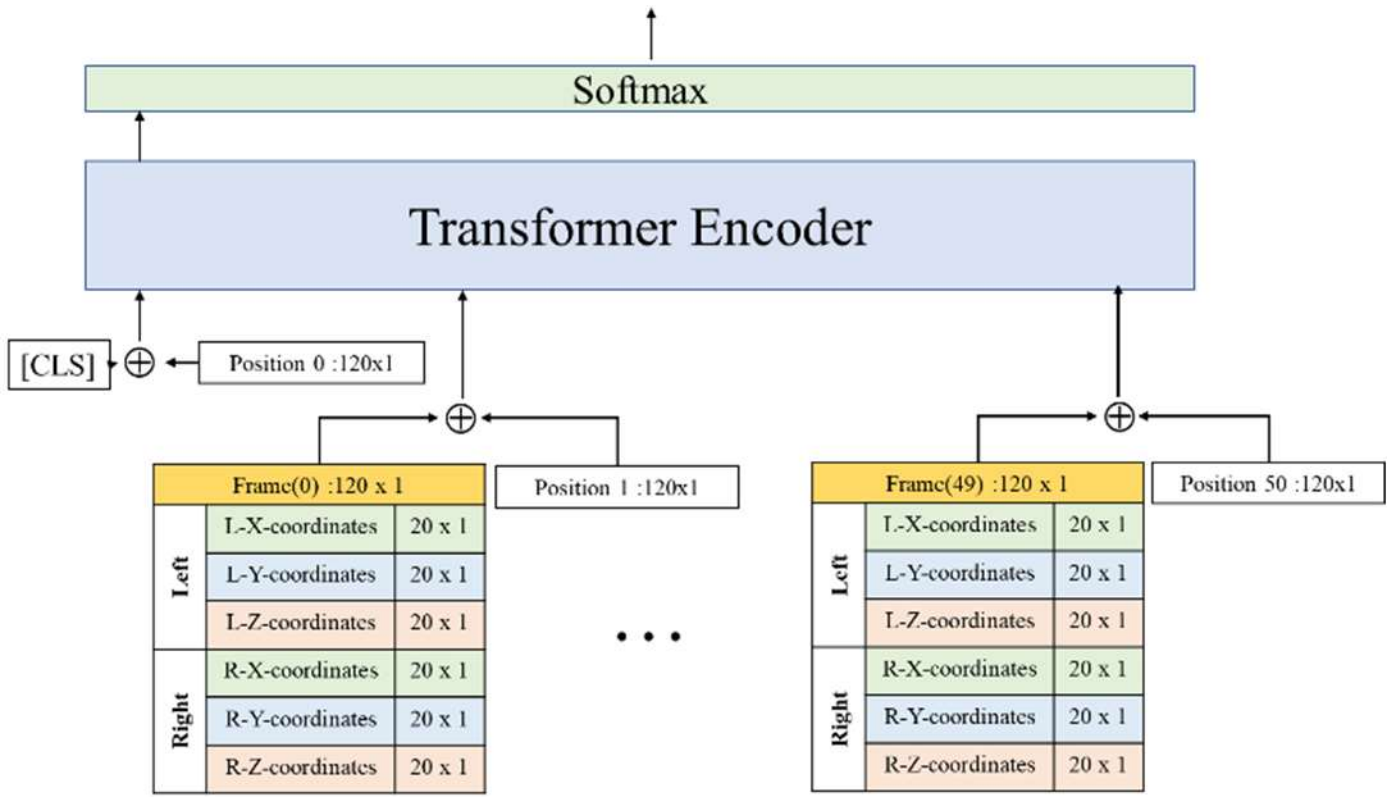

Fig. 2: An overview of the proposed model for ISLR.

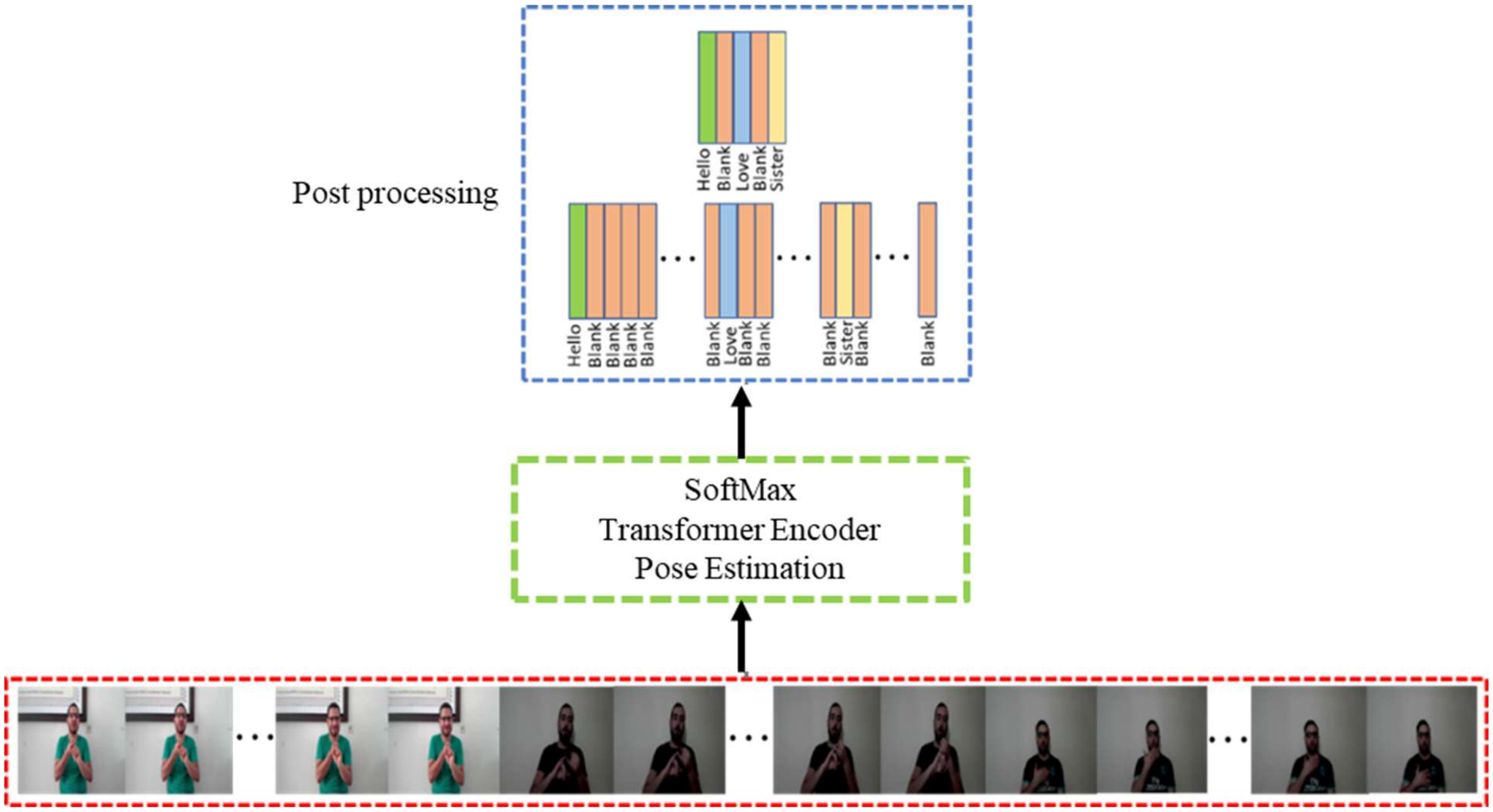

Fig. 3: An overview of the proposed model for CSLR.

### 3.1. The proposed model architecture

Here, the architecture of the proposed model is introduced in detail. As Fig. 1 and 2 show, the proposed model contains the main steps as follows:

- **3D hand keypoints estimation**: The OpenPose library is used to obtain 21 3D hand keypoints. After the normalization, we only use 20 3D keypoints in the proposed model.
- **Features enrichment**: Relying on the Transformer capabilities, it gets the 3D hand keypoints corresponding to each frame and enriches them using the self-attention mechanism. The features of all frames in an isolated sign video are fed to the classification layer in the next step.
- **Classification**: Using the SoftMax layer, the occurrence probabilities of all classes are obtained in this step. The maximum probability will use to identify the final label of the input sign video.
- **Post-Processing:** This step uses the post-processing algorithm borrowed from the [3] to detect the isolated signs in a continuous sign sequence. This step is applied only to the continuous sign videos. More details of this step are available in [3].

### 3.2. Transformer model

Considering the problem definition in this paper, we have a continuous sign video CS = ($F_1$, $F_2$, …, $F_N$) with N frames and a target label T = ($S_1$, $S_2$, …, $S_M$) with M labels corresponding to the M isolated sings. The proposed model aims to recognize the isolated signs using the Transformer model as well as the post-processing algorithm. To this end, the 3D hand keypoints $P_{in}^{i} = \left(P_{in}^{i1}, P_{in}^{i2}, \ldots, P_{in}^{iK}\right)$ are input to the Transformer model to obtain the enriched features $P_{out}^{i} = \left(P_{out}^{i1}, P_{out}^{i2}, \ldots, P_{out}^{iK}\right)$. In this way, the input features are encoded and added to the positional encoding [27] to preserve the order of the frames in the input video:

$$P_{out}^{ij} = End\left(P_{in}^{ij}\right), \quad (1)$$

which the $P_{in}^{ij}$ and $P_{out}^{ij}$ are the jth frame of the ith isolated sign video and the corresponding encoded vector. The n similar layers, including the Multi-Head Attention (Multi-HA) and Feed-Forward (FF) sub-layers in each layer, are used in the Encoder. A Multi-HA makes a weighted contextual representation using the multiple projections of the scaled dot-production attention, which is defined as a linear combination of values V, queries Q, keys K, and dimensionality $d_k$ :

$$Att(Q, K, V) = \text{Somax}\left(\frac{QK^T}{\sqrt{d_k}}\right)V, \quad (2)$$

In addition to this, two linear layers with the ReLU activation function are used in the FF sub-layer. The output of the Transformer model is fed to a FC layer equipped with a Softmax activation function for last classification:

$$C^i = \text{Softmax}\left(P_{out}^{ij}\right), \quad (3)$$

where $C^i$ is a vector including the probabilities of all classes. The post-processing algorithm uses this vector, $C^i$, to determine the final labels of the continuous sign video. It is worth mentioning that using the sliding window approach in the post-processing algorithm, a predefined number of the frames in a continuous sign video is fed to the Transformer model, which is already trained on the isolated signs. Using the post-processing algorithm, the final labels of a continuous sign video will be obtained as follows:

$$\text{Labels} = \text{Post-Process}\left(C^i\right). \quad (4)$$

where $\text{Labels}$ and $\text{Post-Process}$ are the vector of the final labels and the post-processing algorithm, respectively.

# 4. Results

Here, the details of the datasets as well as the experimental results are presented.

## 4.1. Datasets

Due to the need for both continuous sign videos as well as the corresponding isolated signs in each continuous sign video, we make such a dataset using two datasets including the isolated sign videos. To this end, RKS-PERSIANSIGN [28] and ASLLVD [29] datasets, containing the 100 Persian and American sign words, are employed to create the continuous sign videos. While the isolated sign videos are pre-processed during the training phase to have a similar frame number per all videos, we do not apply any pre-processing to the test data and employ them with different frame numbers. Finally, we will have 100 and 7 continuous sign videos, including 100 Persian and American isolated signs with different frame numbers, in the RKS-PERSIANSIGN and ASLLVD datasets, respectively.

## 4.2. Implementation details

The Python software with NVIDIA Tesla K80 GPU, and the PyTorch library are employed in our implementation. The Adam optimizer with a mini-batch size of 50, a learning rate of 0.005, dividing by 10 every 10 epochs, 200 epochs with early stopping, a weight decay of 1e-4, and a momentum of 0.92 are used in the model. Two datasets, RKS-PERSIANSIGN and ASLLVD, randomly divided into training (80%) and testing (20%) sets, are employed for the model evaluation.

## 4.3. Experimental results

In this sub-section, we make an ablation study on the proposed model for both isolated and continuous sign videos as follows:

- **Isolated signs**: During the model training, a Transformer model with 12 layers and 8 heads is used with the 3D hand skeleton inputs. The concatenated outputs of the Transformer model corresponding to all frames of an isolated sign video are fed to a FC layer with 100 neurons number for the final sign classification. Results of the proposed model with different layer numbers as well as the head number on two datasets are shown in Table 1. As this table shows, the best result is obtained using 12 layers number with 8 heads.
- **Continuous signs:** Here, we do not have a training phase. The trained model on the isolated signs is used to detect the isolated signs in continuous sign videos created using two datasets, RKS-PERSAINSIGN and ASLLVD, as we discussed in sub-section 4.1. Using the average frame numbers of the isolated sign videos, we used 50 frames for each isolated sign video during the training phase. A sliding window, including 50 frames in each window, is applied to the continuous sign video. Relying on the post-processing algorithm, the only first occurrence of the consecutive occurrence of an isolated sign, with a Softmax probability higher than the 0.51 threshold, is considered in the final label of a continuous sign video [3]. The results of this step have been shown in Table 2 and Table 3. The results of these tables show that the proposed model has a total of 9 and 8 false recognition numbers for 100 and 7 videos of the RKS-PERSIANSIGN and ASLLVD datasets, respectively. Each video includes 100 isolated signs in each dataset.
  To show the effectiveness of the post-processing approach, the results of the average of Softmax outputs as well as the number of false recognitions on two datasets for boundary detection in continuous sign videos have been shown in Table 4 and 5. Relying on the post-processing approach, the proposed model is more accurate than the other models in these tables.

*Table 1: Results of the ISLR on two datasets: RKS-PERSIANSIGN and ASLLVD.*

| Model | RKS-PERSAIN | ASLLVD |
|---|---|---|
| 1 layer with 4 heads | 76.05 | 73.00 |
| 1 layer with 8 heads | 78.25 | 74.05 |
| 1 layer with 12 heads | 80.05 | 75.20 |
| 2 layers with 4 heads | 80.60 | 76.15 |
| 2 layers with 8 heads | 82.05 | 77.00 |
| 2 layers with 12 heads | 83.00 | 78.05 |
| 3 layers with 4 heads | 84.65 | 81.10 |
| 3 layers with 8 heads | 85.80 | 82.00 |
| 3 layers with 12 heads | 87.30 | 83.25 |
| 4 layers with 4 heads | 90.90 | 86.20 |
| 4 layers with 8 heads | 91.40 | 87.10 |
| 4 layers with 12 heads | 92.00 | 88.00 |
| 8 layers with 4 heads | 92.60 | 91.05 |
| 8 layers with 8 heads | 94.20 | 91.60 |
| 8 layers with 12 heads | 94.75 | 92.20 |
| 12 layers with 4 heads | 98.70 | 93.00 |
| 12 layers with 8 heads | **99.80** | **95.80** |
| 12 layers with 12 heads | 98.80 | 93.50 |

*Table 2: Results on the ASLLVD dataset for CSLR.*

| Concatenated sign video | Avg of Softmax output of Recognized class | Ground truth Word class | Softmax output of Ground truth Word class | Recognized class | Softmax output of Recognized class |
|---|---|---|---|---|---|
| 1 | **0.68** | **63** | **0.35** | **45** | **0.36** |
| 2 | **0.69** | **8** | **0.34** | **18** | **0.38** |
| 3 | **0.68** | **63** | **0.44** | **45** | **0.47** |
|  | **0.68** | **8** | **0.35** | **18** | **0.36** |
| 4 | **0.68** | **64** | **0.32** | **51** | **0.34** |
| 5 | **0.67** | **18** | **0.30** | **8** | **0.35** |
| 6 | **0.68** | **63** | **0.30** | **45** | **0.33** |
| 7 | **0.69** | **45** | **0.33** | **63** | **0.34** |

*Table 3: Results on the RKS-PERSIANSIGN dataset for CSLR.*

| Concatenated sign video | Avg of Softmax output of Recognized class | Ground truth Word class | Softmax output of Ground truth Word class | Recognized class | Softmax output of Recognized class | Concatenated sign video | Avg of Softmax output of Recognized class | Ground truth Word class | Softmax output of Ground truth Word class | Recognized class | Softmax output of Recognized class | Concatenated sign video | Avg of Softmax output of Recognized class | Ground truth Word class | Softmax output of Ground truth Word class | Recognized class | Softmax output of Recognized class |
|---|---|---|---|---|---|---|---|---|---|---|---|---|---|---|---|---|---|
| 1 | 0.99 | **17** | **0.43** | **19** | **0.46** | 35 | 0.99 | - | - | - | - | **69** | 0.98 | **63** | **0.43** | **45** | **0.45** |
| 2 | 0.98 | - | - | - | - | 36 | 0.99 | - | - | - | - | **70** | 0.98 | - | - | - | - |
| 3 | 0.98 | - | - | - | - | 37 | 0.98 | **66** | **0.40** | **86** | **0.42** | **71** | 0.99 | - | - | - | - |
| 4 | 0.98 | - | - | - | - | 38 | 0.98 | - | - | - | - | **72** | 0.99 | - | - | - | - |
| 5 | 0.99 | - | - | - | - | 39 | 0.99 | - | - | - | - | **73** | 0.99 | - | - | - | - |
| 6 | 0.99 | - | - | - | - | 40 | 0.99 | - | - | - | - | **74** | 0.98 | - | - | - | - |
| 7 | 0.99 | - | - | - | - | 41 | 0.99 | - | - | - | - | **75** | 0.99 | - | - | - | - |
| 8 | 0.99 | - | - | - | - | 42 | 0.99 | - | - | - | - | **76** | 0.99 | - | - | - | - |
| 9 | 0.99 | - | - | - | - | 43 | 0.99 | - | - | - | - | **77** | 0.99 | **63** | **0.43** | **45** | **0.44** |
| 10 | 0.99 | **17** | **0.44** | **19** | **0.45** | 44 | 0.99 | - | - | - | - | **78** | 0.99 | - | - | - | - |
| 11 | 0.97 | - | - | - | - | 45 | 0.98 | - | - | - | - | **79** | 0.99 | - | - | - | - |
| 12 | 0.98 | - | - | - | - | 46 | 0.99 | **45** | **0.43** | **63** | **0.44** | **80** | 0.99 | - | - | - | - |
| 13 | 0.99 | - | - | - | - | 47 | 0.99 | - | - | - | - | **81** | 0.98 | - | - | - | - |
| 14 | 0.99 | - | - | - | - | 48 | 0.99 | - | - | - | - | **82** | 0.99 | - | - | - | - |
| 15 | 0.99 | - | - | - | - | 49 | 0.99 | - | - | - | - | **83** | 0.99 | - | - | - | - |
| 16 | 0.99 | - | - | - | - | 50 | 0.99 | - | - | - | - | **84** | 0.99 | - | - | - | - |
| 17 | 0.98 | - | - | - | - | 51 | 0.98 | - | - | - | - | **85** | 0.99 | - | - | - | - |
| 18 | 0.99 | - | - | - | - | 52 | 0.99 | - | - | - | - | **86** | 0.99 | - | - | - | - |
| 19 | 0.99 | - | - | - | - | 53 | 0.99 | - | - | - | - | **87** | 0.98 | - | - | - | - |
| 20 | 0.99 | - | - | - | - | 54 | 0.99 | - | - | - | - | **88** | 0.99 | - | - | - | - |
| 21 | 0.99 | - | - | - | - | 55 | 0.99 | - | - | - | - | **89** | 0.99 | - | - | - | - |
| 22 | 0.99 | - | - | - | - | 56 | 0.99 | - | - | - | - | **90** | 0.99 | - | - | - | - |
| 23 | 0.99 | - | - | - | - | 57 | 0.99 | - | - | - | - | **91** | 0.99 | - | - | - | - |
| 24 | 0.99 | - | - | - | - | 58 | 0.99 | - | - | - | - | **92** | 0.99 | - | - | - | - |
| 25 | 0.99 | - | - | - | - | 59 | 0.99 | **17** | **0.44** | **19** | **0.45** | **93** | 0.99 | - | - | - | - |
| 26 | 0.99 | **19** | **0.44** | **17** | **0.46** | 60 | 0.99 | - | - | - | - | **94** | 0.99 | **45** | **0.43** | **63** | **0.44** |
| 27 | 0.96 | - | - | - | - | 61 | 0.97 | - | - | - | - | **95** | 0.99 | - | - | - | - |
| 28 | 0.99 | - | - | - | - | 62 | 0.98 | - | - | - | - | **96** | 0.99 | - | - | - | - |
| 29 | 0.99 | - | - | - | - | 63 | 0.99 | - | - | - | - | **97** | 0.98 | - | - | - | - |
| 30 | 0.99 | - | - | - | - | 64 | 0.99 | - | - | - | - | **98** | 0.99 | - | - | - | - |
| 31 | 0.99 | - | - | - | - | 65 | 0.99 | - | - | - | - | **99** | 0.99 | - | - | - | - |
| 32 | 0.99 | - | - | - | - | 66 | 0.99 | - | - | - | - | **100** | 0.99 | - | - | - | - |
| 33 | 0.99 | - | - | - | - | 67 | 0.99 | - | - | - | - |  |  |  |  |  |  |
| 34 | 0.99 | - | - | - | - | 68 | 0.99 | - | - | - | - |  |  |  |  |  |  |

*Table 4: Ablation analysis of the average of Softmax outputs on two datasets for boundary detection in continuous sign videos: The impact of post-processing.*

| Model | Dataset | |
|---|---|---|
| | RKS-PERSIANSIGN | ASLLVD |
| Proposed (Transformer without post-processing) | 0.4260 | 0.3350 |
| Proposed (Transformer with Post-Processing) | 0.9930 | 0.6820 |

*Table 5: Ablation analysis of the number of false recognitions on two datasets for boundary detection in continuous sign videos: The impact of post-processing.*

| Model | Dataset | |
|---|---|---|
| | RKS-PERSIANSIGN | ASLLVD |
| Proposed (Transformer without post-processing) | 615 | 105 |
| Proposed (Transformer with Post-Processing) | 9 | 8 |

- **Comparison with same tasks:** After preparing the continuous video samples similar to [3], there are a total of 100 and 7 continuous video samples in the RKS-PERSIANSIGN and ASLLVD datasets, respectively. Each video includes 100 isolated signs. Using the prepared data, the proposed model is compared to similar tasks in [3, 25] (Table 6 and Table 7). Relying on the Transformer capabilities, the proposed model has a comparative accuracy with the suggested model in [25] and a higher accuracy than the model in [3]. Using the self-attention mechanism and parallel computing, the proposed model obtains richer features in the data that lead to an accurate model. However, there are still some false recognition samples in the approximately identical signs with many common frames. Using the additional samples and features in the false recognized signs could help to reduce the false classifications occurred in a low inter-class variation.

*Table 6: Comparison of the average of Softmax outputs on two datasets for CSLR.*

| Model | Dataset | |
|---|---|---|
| | RKS-PERSIANSIGN | ASLLVD |
| SVD + Post-Processing [3] | 0.9800 | 0.5900 |
| GCN + Post-Processing [25] | 0.9945 | 0.6835 |
| Proposed (Transformer + Post-Processing) | 0.9930 | 0.6820 |

*Table 7: Comparison of the number of false recognitions on two datasets for CSLR.*

| Model | Dataset | |
|---|---|---|
| | RKS-PERSIANSIGN | ASLLVD |
| SVD + Post-Processing [3] | 24 | 12 |
| GCN + Post-Processing [25] | 9 | 7 |
| Proposed (Transformer + Post-Processing) | 9 | 8 |

## 4.4. Discussion

In this paper, we propose a Transformer model comprising 12 layers, 8 heads, and designed for processing 3D hand skeleton inputs. The model is exclusively trained on isolated signs and subsequently employed during the inference phase on continuous sign videos generated from two datasets, namely RKS-PERSIAN SIGN and ASLLVD. The results obtained demonstrate promising outcomes on both the RKS-PERSIAN SIGN and ASLLVD datasets. However, several challenges merit discussion. One notable challenge pertains to the dataset. In the absence of a dataset encompassing both continuous sentences and their corresponding isolated signs, we artificially constructed such a dataset. It is acknowledged that the synthetic nature of this dataset compromises its realism, thereby restraining our ability to make definitive claims about the generalization of our results. Consequently, it is imperative to assess our model on a real-world dataset featuring both continuous sentences and their associated isolated signs. Subsequent to this evaluation, a comprehensive discussion on the generalization capabilities of the proposed model can ensue. Another challenge pertains to the observation that distinct users execute signs of varying durations, leading to disparate frame lengths. In the realm of ISLR, the commonly employed padding mechanism in NLP tasks to accommodate variable sentence lengths can be utilized to manage the variability in video frame numbers. However, due to the specific focus of this paper, a predetermined number is essential for detecting sign boundaries using a sliding window approach. Accordingly, we opt for the average length of all signs as the predefined value in this context. It is noteworthy that the mechanism for determining this predetermined number may yield varying results.

# 5. Conclusion and future direction

In this paper, we have three contributions, which are discussed as follows:

- **Model:** Relying on the post-processing method presented in [3] for boundary detection of the isolated signs in a continuous sign video [3], this paper aimed to improve the recognition accuracy by presenting a new model using the Transformer model. Using the self-attention mechanism embedded in the Transformer model as well as the parallel computing of this model, the proposed model tackles the challenge of boundary detection of the isolated signs in the continuous video stream. Our different analysis of the proposed model confirmed the necessity as well as the efficiency of the model. Analysis of the different head and layer numbers in the Transformer model showed that the highest performance is obtained using 12 and 8 layer and head numbers, respectively. While the proposed model is trained on isolated signs, we do not have a training phase for continuous videos. The trained model on the isolated signs is used to detect the isolated signs in continuous sign videos created using two datasets, RKS-PERSAINSIGN and ASLLVD. Using the sliding window approach, the value of 0.51 is considered as a default value for accepting or rejecting a recognized isolated sign in the current sliding window. The intuition behind this assumption about the default value refers to the Softmax property that calculates the output probabilities of all classes summed to 1. So, only one class above this default value can exist in the current sliding window, including the output class probabilities. Considering the "Blank" label for false classes with the Softmax probabilities lower than the default value, these classes are suppressed little by little. In this way, we will have the" Blank" labels in the output instead of the false recognized labels. The average Softmax outputs of the proposed model on the RKS-PERSIANSIGN and ASLLVD datasets are of 0.99 and 0.68, respectively. As we expected, relying on the higher sample number per each class in the RKS-PERSIANSIGN dataset as well as the generally better efficiency of the Deep Learning-based models in coping with the larger amount of data, the final recognition accuracy as well as the average Softmax probability of this dataset is higher than the same criteria in the ASLLVD dataset.
- **Multi-tasking:** Using the trained model on the isolated sign videos, it is used for continuous sign videos. So, both of the ISLR and CSLR are handled by only one model. More concretely, the proposed model for boundary detection in CSLR includes two sub-models: a model for ISLR and a post-processing approach. The results of the model for both ISLR and CSLR have been reported and discussed.
- **Performance:** Aiming to improve the model performance, we proposed a model using a Transformer model. After preparing the continuous video streams using the isolated signs of the two datasets, the proposed model has been analyzed and compared with the same tasks. Furthermore, the impact of the post-processing approach has been reported and discussed. Benefiting from the post-processing methodology in [3] as well as the self-attention capabilities embedded in the Transformer model, the proposed model presents promising results compared to the same models. However, there are some false recognition samples in the proposed model that need to be solved. In future work, we will focus on real-world continuous sign videos equipped with the corresponding isolated sign videos to analyze the model performance in a real situation. To this end, we would like to prepare this data and present it to the research community.

## DECLARATIONS

### Funding

This research did not receive any specific grant from funding agencies in the public, commercial, or not-for-profit sectors.

### Declaration of Competing Interest

The authors certify that they have no conflict of interest.

### Ethics approval (include appropriate approvals or waivers)

’Not applicable’

### Consent for publication

All authors confirm their consent for publication.

### Availability of data and material (data transparency)

’Not applicable’

### Code availability (software application or custom code

’Not applicable’